\pdfoutput=1

\documentclass[11pt]{article}

\usepackage[]{acl}

\usepackage{times}
\usepackage{latexsym}

\usepackage[T1]{fontenc}

\usepackage[utf8]{inputenc}

\usepackage{microtype}

\usepackage{inconsolata}

\usepackage{enumitem}
\usepackage{multirow}
\usepackage{booktabs}
\usepackage{amsmath}
\usepackage{inconsolata}
\usepackage{stfloats} 
\usepackage{graphicx}
\usepackage{microtype}

\title{AMAQ: Adaptive Mixed-bit Activation Quantization for Collaborative \\ Parameter Efficient Fine-tuning}

\author{
Yurun Song \\
UC Irvine \\
\texttt{yuruns@uci.edu} \\\And
Zhuoyi Yang\\
UC Irvine \\
\texttt{zhuoyy1@uci.edu} \\
\AND
Ian G. Harris \\
UC Irvine \\
\texttt{harris@ics.uci.edu} \\\And
Sangeetha Abdu Jyothi \\
UC Irvine, VMware Research \\
\texttt{sangeetha.aj@uci.edu} \\
}

\begin{document}
\maketitle

\begin{abstract}
Large Language Models (LLMs) are scaling rapidly, creating significant challenges for collaborative server-client distributed training, particularly in terms of communication efficiency and computational overheads. To address these challenges, we implement Parameter-efficient Split Learning, which effectively balances efficiency and performance for collaborative training on low-resource devices. 


To reduce communication overhead in collaborative training, we introduce Adaptive Mixed-bit Activation Quantization (AMAQ), a strategy that progressively compresses activations and gradients from high precision (6–8 bits) to low precision (3–4 bits). AMAQ achieves this by effectively allocating bit budgets across channels based on feature-wise and layer-wise importance using bit regularization.

Under the same bit budgets, AMAQ outperforms fixed-precision approaches, delivering about 2.5\% higher generation accuracy and about 1.3\% better classification accuracy for models like LLaMA3-8B and Qwen2.5-7B. In addition, it significantly enhances training stability and reducing ultra-low bit representation collapse during the training.

Experiments demonstrate that AMAQ integrates effectively into practical multi-machine collaborative training setups, offering superior inference accuracy with only a modest communication overhead for bits adaptation during training. This trade-off makes AMAQ a practical and effective solution for collaborative training with minimal communication cost.

\end{abstract}


\section{Introduction}
LLMs have seen exponential growth in size, reaching hundreds of billions of parameters. Although this scaling yields remarkable performance in natural language processing, it also requires heavy computational and memory resources, making local deployment challenging. At the same time, users increasingly prefer to keep models local, yet full-scale models often exceed the capacity of local hardware. Previous works on client-server training~\cite{gao2024dloradistributedparameterefficientfinetuning, lin2024splitlorasplitparameterefficientfinetuning} attempt to solve these challenges through various strategies, targeting two major issues: 
(i) significant computational demands on client devices, and (ii) high communication overhead due to frequent activation and gradient exchange.

To alleviate the \textit{computational overhead}, parameter-efficient fine-tuning (PEFT) techniques have been introduced~\cite{li2021prefixtuningoptimizingcontinuousprompts, liu2022ptuningv2prompttuning, hu2021loralowrankadaptationlarge}, allowing only a subset of parameters to be updated, rather than the entire model's weights. However, even with PEFT, the size and complexity of modern LLMs often exceed the capabilities of client devices. 
This challenge led to the adoption of split learning~\cite{thapa2022splitfed, lin2024splitlorasplitparameterefficientfinetuning}, where the model is partitioned between the client and a remote server, enabling partial on-device processing. While split learning reduces local resource demand, it introduces a new challenge --- high communication overhead to transmit intermediate activations over networks.
To efficiently tackle these two challenges--- computational overhead and communication overhead---we implement a unified framework for Precision-adaptive, collaborative Parameter-efficient Split Learning. 

Low-precision training and activation compression serve as promising strategies to reduce both memory usage and communication overhead. Methods employing FP8~\cite{deepseekai2025deepseekv3technicalreport} or FP4~\cite{wang2025optimizinglargelanguagemodel} have demonstrated that using fewer bits can significantly accelerate training. Although activation compression has been extensively studied in computer vision, its application in Large Language Models remains relatively underexplored recently. Recent works, such as Activation-aware Weight Quantization (AWQ)~\cite{lin2024awqactivationawareweightquantization}, smoothQuant~\cite{xiao2024smoothquantaccurateefficientposttraining}, Channel-Wise Quant~\cite{chen2024channel}, and Group-Wise Quant~\cite{yang2024gwq}, aim to preserve performance while minimizing weight quantization errors in LLMs instead of focusing on activations. These works didn't study weight quantization for low precision (4 bits and under).


We propose Adaptive Mixed-bit Activation Quantization (AMAQ), which adaptively assigns higher precision (6-8 bits) to critical features and lower precision (3-4 bits) to less important ones. This quantization is applied to all activations, gradients, and transmitted modules during training, reducing communication overhead while preserving model performance.

Building upon activation compression strategies such as AQ-SGD \cite{wang2023finetuninglanguagemodelsslow}, Learned Gradient Linear Symmetric Quantization (LG-LSQ) \cite{lin2022lglsqlearnedgradientlinear}, and Rectified Straight Through Estimator (Re-STE) \cite{wu2023estimatormeetsequilibriumperspective}, our approach employs a gradual quantization schedule to decrease precision from higher to lower bits. Overall, our framework enables split learning to reduce computational and communication costs while achieve high performance in distributed LLM training. Our contributions fall into three categories:

\begin{figure*}[!ht]
  \centering 
  \includegraphics[width=0.95\linewidth]{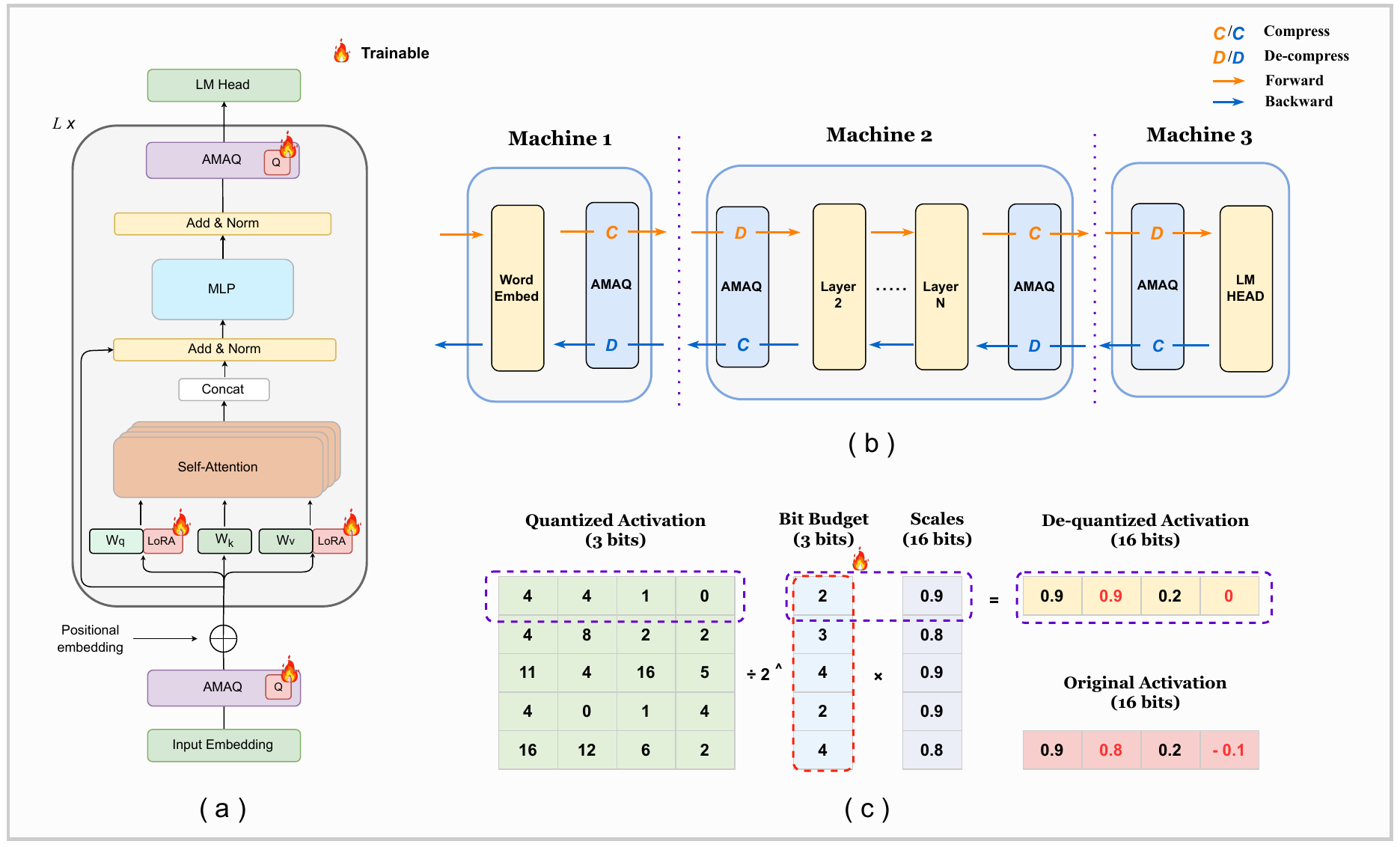}

  \caption{(a) Activation quantization pipeline in our framework, where both LoRA and the learnable quantization parameters (Q) are jointly optimized during training.
(b) AMAQ deployment in a split learning setup, enabling model parallelism across multiple machines. In this setting, the model can be partitioned, and activations are compressed and decompressed using the AMAQ quantizer for efficient network communication.
(c) Visualization of dequantized activations using AMAQ. Unlike prior approaches, our method incorporates a trainable bit budget controlled by a bit-regularization, allowing dynamic control over the bit allocation of each channel during training.}

  \label{fig:use_case}
\end{figure*}

\begin{itemize}

    \item \textbf{Adaptive Activation Quantization (\textbf{AMAQ}) for Training} 
    We propose an Adaptive Mixed-bit activation quantization strategy that adaptively assigns bits based on feature importance, effectively reducing communication overhead and accelerating both training and inference.

    \item \textbf{Real-World Evaluation of AMAQ's Communication Efficiency}
    We deploy AMAQ in distributed real-world settings to measure its impact on GPU usage, communication latency, and activation transmission size to validate its efficiency under practical split learning environments.
    
    \item \textbf{Investigation of Low-Bit Activation Quantization for different LLMs} 
    Our work explores how different LLMs handle activation quantization at extremely low bit widths (e.g., 4 bits, and 3 bits) on both classification and generation tasks.


\end{itemize}

\section{Related Work}
\textit{Quantization} is widely used to reduce the memory and computational footprint of neural networks, especially on edge devices. Broadly, quantization methods fall into two categories: Post-Training Quantization (PTQ) and Quantization-Aware Training (QAT). PTQ methods~\cite{Min-Max-Quantization,KL-Divergence,Bias-Correction} apply quantization to a pre-trained full-precision model without retraining. 
In contrast, QAT methods~\cite{lin2022lglsqlearnedgradientlinear,wu2023estimatormeetsequilibriumperspective,wang2023finetuninglanguagemodelsslow} simulate quantization during training to enable the model to adapt to quantization noise. 

\textit{Sparsity} is an alternative compression technique, involving a variety of methods such as sparsifying the gradient of activations~\cite{meProp} and sparsifying softmax attention~\cite{sparsesinkhornattention}. Dynamic \textit{pruning} is another popular method for compression, involving dynamically pruning and regrowing connections~\cite{rigL}. However, these approaches often require specialized training regimes or custom hardware support.



\textit{PEFT} methods, a practical alternative to full model fine-tuning,  introduce small, trainable modules or embeddings while keeping the base model frozen, significantly reducing computational and memory costs. Among the most prominent methods are LoRA~\cite{hu2021loralowrankadaptationlarge}, which inserts trainable low-rank matrices into attention layers; adapter modules~\cite{adaptor-modules}, which add lightweight neural networks between transformer layers; and prompt-based approaches such as Prefix-Tuning \cite{prefix-tuning} and prompt tuning \cite{liu2022ptuningv2prompttuning}, which optimize task-specific input representations. 
These strategies offer high performance across a variety of NLP tasks, including text classification and question answering, making them particularly appealing in resource-constrained settings.

Recent work has shown that combining PEFT methods with \textit{Differential Privacy} (DP) can preserve utility with minimal performance degradation. LoRA notably improves speed and memory efficiency with DP-SGD~\cite{yu2022DP}. Subsequent DP-PEFT frameworks~\cite{li2022large} introduce DP-Adam and Ghost Clipping to reduce memory overhead further. RAPT ~\cite{li2023privacy} extends DP-PEFT to prompt tuning via API-based models that mitigate performance drop in privacy-sensitive settings. These results highlight PEFT with DP as a practical solution for private LLM fine-tuning.

\section{Methodology}

\subsection{Adaptive Mixed-bit Activation Quantization}

Quantization-aware training (QAT) offers high performance at higher bit-widths. However, pushing QAT to extremely low bit-widths (fewer than 4 bits) remains challenging. Although recent works have achieved high performance with low-bit QAT on weights~\cite{lin2022lglsqlearnedgradientlinear}, the compression of activations in low bits continues to be a bottleneck~\cite{choi2018pactparameterizedclippingactivation,zhou2018dorefanettraininglowbitwidth}.

Rather than directly quantizing activations to extremely low bit-widths, our technique, \emph{Adaptive Mixed-bit Activation Quantization} (AMAQ), dynamically transitions from a higher bit-width setting to a lower bit-width as the training progresses, as demonstrated in Figure~\ref{fig:use_case} (c). This achieves higher accuracy than fixed-bits approaches as our experiments prove. Our approach introduces additional trainable parameters, gating parameters \(Q\), which are optimized separately from the model weights and LoRA weights. We treat these gating parameters as an independent group in the optimizer, allowing us to control the convergence speed of AMAQ. We explicitly regularize \(Q\) to steer the quantization process. We define the effective bit-width as:
\begin{figure}[ht]
  \centering 
\includegraphics[width=0.95\linewidth]{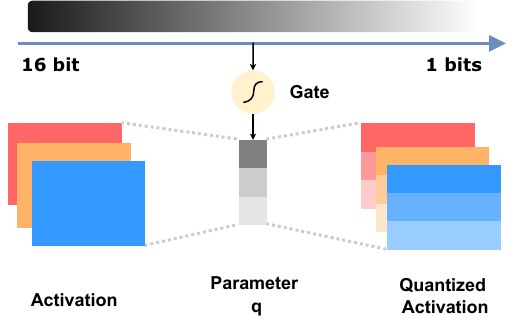}
\caption{
  Adaptive Mixed-bit Activation Quantization employs a learnable parameter to control the bit-width of each activation channel through a gating mechanism.}
  \label{fig:mixbits}
\end{figure}

\[
\text{Bit-width} = \textit{min} +(\textit{max} - \textit{min}) \times \sigma~(~\alpha \cdot Q~)
\]
where \(Q\) is a vector of shape \(1 \times H\) (or alternatively \(1 \times \text{max\_seqLen}\) for per-token quantization) that controls the quantization bit-width for each feature. \(\sigma\) denotes an sigmoid function as a gating mechanism, and \(\alpha\) is a scaling coefficient that is inversely related to the learning rate for \(Q\); a larger \(\alpha\) effectively corporate with a smaller learning rate for stability and smoothens bit-width reduction. $min$ is the minimum bit-width and $max$ is the maximum bit-width. For example, in Figure~\ref{fig:mixbits}, we illustrate a search range, from 1 to 16 bits. The gating mechanism then modulates the non-linear range between these bit configurations. This smooth decrement of bit-width ensures a stable learning trajectory, preventing abrupt losses in representational precision during training while retaining the benefits of high-bit QAT in the early phases. 




Quantization-aware training often contend with non-differentiable operations, such as rounding, which is essential for quantization. To overcome this issue, we apply Straight Through Estimator (STE) to approximate the gradients through these operations, which effectively enables gradient flow during backpropagation.




\subsection{Bits Regularization}
To control the contribution of the gating parameters \(Q\) during training, we apply L2 regularization to \(Q\) and introduce L2 bits loss. The regularization norms are defined as:
\[
\text{Bits\_Loss} = \frac{1}{n}\sum_{i=1}^{n} \sigma(\alpha \cdot q_i)^2
\]
We then combine these regularization terms into our final loss function as follows:
\[
\text{Loss} = \text{QAT\_Loss} + \beta \times \text{Bits\_Loss}
\]
where \(\beta\) is a hyperparameter that balances the QAT loss with the bit adaptation loss. 
While L2 norm regularization is commonly used for training stability, it can be substituted with L1 norm regularization. The benefit of using the L1 norm is that it encourages sparsity in the activation bit allocation, which is advantageous for compression scenarios. In contrast, L2 norm tends to be more stable during quantization-aware training.



However, we find that the model continues to reduce the effective bit-width even after reaching the desired quantization level. To address this issue, we introduce a clipping function that constrains the mean bit-width of the activations once the AMAQ process attains the target bit. 
\[
\scalebox{0.92}{$
  \displaystyle
  \text{Loss}^{\mathrm{clip}}
  = \text{Loss}
  + \frac{\beta}{n}
    \sum_{i=1}^{n}
      \sigma~\!\bigl(\alpha\cdot\,\mathrm{clip}(q_i,q_{\min},q_{\max})\bigr)^{2}
$}
\]
Importantly, instead of clipping the bit-width of any individual activations, we clip the mean of all activations that need to be quantized. This strategy preserves the relative importance of activations. For example, if the input activation is effectively 3.5 bits and the output activation is 4.5 bits, the overall quantized bit-width remains at the desired level of 4 bits after clipping, thus maintaining the relative differences.

\begin{table*}[htbp]
\centering
\scalebox{0.98}{
\begin{tabular}[\textwidth]{cc|c|c|c|c|c}
\toprule
\multicolumn{2}{c|}{\textbf{PPL} $\downarrow$}                           & \textbf{QAT-Tensor} & \textbf{QAT-Channel} & \textbf{QAT-Group} & \textbf{AQ-SGD} & \textbf{AMAQ} \\ 
\midrule
\multicolumn{1}{c|}{\multirow{3}{*}{\textbf{LLaMA3 8B}}}  & GSM8k        &  1.864          &   1.652          &    1.619       &  1.636       &  \textbf{1.614}      \\  
\multicolumn{1}{c|}{}                                      & MATH        &  2.405          &   1.961          &    1.912       &  1.924      &  \textbf{1.905}      \\ 
\multicolumn{1}{c|}{}                                      & Code-Alpaca &  2.068        &    1.810         &     1.752      &  1.805      &  \textbf{1.750}      \\ 
\midrule

\multicolumn{1}{c|}{\multirow{3}{*}{\textbf{Qwen2.5 14B}}} & GSM8k       & 1.956      &   3.615          &   1.602        &   1.477     &   \textbf{1.462}     \\  
\multicolumn{1}{c|}{}                                      & MATH        & 2.106      &   2.019         &    1.793       &   1.694     &   \textbf{1.667}     \\  
\multicolumn{1}{c|}{}                                      & Code-Alpaca & 3.413      &   1.730         &  1.907        &    1.919         &  \textbf{1.609}      \\ 
\midrule

\multicolumn{1}{c|}{\multirow{3}{*}{\textbf{Phi-3 Medium}}}   & GSM8k       & 1.749           &  1.489            &   1.488        &  1.426      &  \textbf{1.407}      \\ 
\multicolumn{1}{c|}{}                                      & MATH        & 2.279           &  1.757           &   1.743        &   1.648     &   \textbf{1.634}     \\  
\multicolumn{1}{c|}{}                                      & Code-Alpaca & 2.102           &  1.605           &   1.588        &   1.736     &   \textbf{1.504}     \\ 
\bottomrule
\end{tabular}}
\caption{Perplexity comparison of 4-bit activation compression techniques for LoRA Training. Compression is applied to input and output activations only.} 
\label{ppl}
\end{table*}



\section{Experiments}

Our experiments are conducted under two environments: single-machine and multi-machine. Single-machine evaluations are used to measure the performance of baseline. In multi-machine environments, we deploy a split learning framework to assess real-world computational efficiency and communication overhead, as described in Figure~\ref{fig:use_case} (b). Specifically, we quantify the memory usage, training time, activation transmission size and inference latency.

We evaluate two distinct setup for AMAQ, as shown in the Figure~\ref{fig:use_case} (a): (1) quantization on only the most challenging layers - activation after input word embeddings and before output LM head, which critically impact model performance (2) full-model quantization across all transformer layers.





We employ L2 normalization in all experiments. We explore various bit-width ranges for activation quantization. In the 4-bit experiments, the bit-width can vary from 1 to 16, starting from an initial bit of 8 bits. For 3-bit quantization, the bit-width ranges from 1 to 8, beginning at 6 bits. 
A clipping function prevents the bit-width from dropping below the target, maintaining fluctuations within approximately $\pm ~ 0.1$ bits of the desired value. For fair comparisons in our experiments, we set $\alpha$ = 1.




To demonstrate its general applicability, we evaluate our approach across multiple LLMs: LLaMA3-8B, Qwen2.5 (7B and 14B), and Phi-3-Medium. We assess model performance using three key metrics: perplexity (PPL) , Exact Match (for mathematical reasoning), and Pass@1 (for code generation capability).
Additional experiments includes full-layer activation quantization, intermediate-layer full fine-tuning, and other ablation studies.





Our distributed experiments are implemented across two servers, with A6000 and RTX 3090 GPUs, respectively, interconnected by a Wi-Fi network to simulate a real world case. Our experiments use the PyTorch distributed package.

\begin{figure*}[h!]
  \centering
 \includegraphics[width=\linewidth]{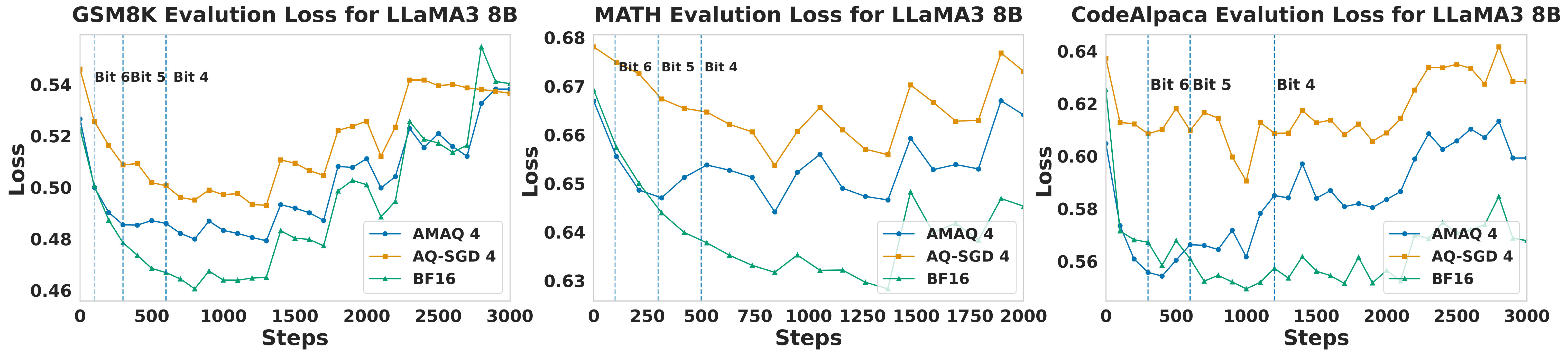}
 \caption{Evaluation of loss performance on GSM8K, MATH, and CodeAlpaca benchmarks for AMAQ, comparing BF16 and AQ‑SGD across various bit-width for both input and output activation quantization.} \label{plot}
\end{figure*}

\subsection{Datasets}
We leverage a diverse suite of datasets to assess model performance across reasoning, coding, mathematics, and various question-answering tasks. We use BoolQ \cite{clark2019boolqexploringsurprisingdifficulty} for classification tasks, ARC-C \cite{allenai:arc} for reading comprehension, and  Winogrande \cite{sakaguchi2019winograndeadversarialwinogradschema} and CommonSenseQA \cite{talmor-etal-2019-commonsenseqa} to evaluate commonsense reasoning. We employ MATH \cite{hendrycksmath2021} and GSM8K \cite{cobbe2021trainingverifierssolvemath} for mathematical reasoning, and CodeAlpaca \cite{codealpaca} and HumanEval \cite{chen2021codex} for code generation.

\subsection{Baselines}
\label{subsec:baselines}
\textbf{Tensor-Wise}: This approach applies a single quantization scale to the entire tensor, simplifying computations but potentially overlooking variations within the tensor. \\
\textbf{Channel-Wise}~\cite{chen2024channel}: Each channel of a tensor is assigned its quantization level, allowing for finer granularity and improved accuracy by accommodating inter-channel differences. \\
\textbf{Group-Wise}~\cite{yang2024gwq}: It divides a tensor into groups of channels, assigning a unique quantization level to each group, balancing between the simplicity of tensor-wise and the precision of channel-wise quantization.\footnote{https://github.com/neuralmagic/compressed-tensors/}\\
\textbf{Activation Quantization Stochastic Gradient Descent (AQ-SGD)}~\cite{wang2023finetuninglanguagemodelsslow}: It compresses the changes in activations rather than the activations themselves, enhancing communication efficiency in distributed training without compromising convergence. \\

\begin{table*}[!ht]
\centering
\scalebox{0.82}{
\begin{tabular}{cc|c|c|c|c|c|c|c}
\toprule
 \textbf{LLaMA3-8B}                 & \textbf{Bits}  & \textbf{Boolq}       & \textbf{ARC-C} &  \textbf{Winogrande} & \textbf{CommonSenseQA} & \textbf{GSM8K} & \textbf{MATH} & \textbf{HumanEval} \\
\midrule
\textbf{Few / Zero Shot}              & 16          &  75.7                & 78.6         & 76.1                & 72.6                 &  45.33         &   17.36      &   31.09        \\
\textbf{LoRA}                         & 16          &  89.75                & 80.94              &    84.84                  & 79.60                &  54.66         &  18.98       &   41.46       \\\hline
\midrule
\textbf{AQ-SGD 4}                       & 4           &  89.32                & 79.82        &  83.74              & 79.77                &  52.54         &  17.08       &   32.32         \\ 
\textbf{AMAQ 4}                     & 4 ± 0.1     &  \textbf{89.84}       & \textbf{80.85}        & \textbf{85.31}               & \textbf{80.09}                &  \textbf{53.60}        &  \textbf{17.84}        &   \textbf{37.80}         \\\hline
\midrule
\textbf{AQ-SGD 3}                       & 3           &  89.54                &  76.99       &83.50               & 78.13                 & 45.87         &    14.82      &   30.49    \\ 

\textbf{AMAQ 3}                     &  3 ± 0.1    &  \textbf{89.75}       & \textbf{80.68}        & \textbf{84.13}               & \textbf{79.03}                & \textbf{50.27}         &   \textbf{15.10}       &    \textbf{31.71}       \\\hline

\bottomrule
\end{tabular}}
\caption{LLaMA 3 8B evaluation results using AMAQ for both input and output layer activations. }\label{LLaMA3Performance}
\end{table*}

\begin{table*}[!ht]
\centering
\scalebox{0.85}{
\begin{tabular}{cc|c|c|c|c|c|c|c}
\toprule
 \textbf{Qwen2.5-7B}                 & \textbf{Bits}  & \textbf{Boolq}       & \textbf{ARC-C} &  \textbf{Winogrande} & \textbf{CommonSenseQA} & \textbf{GSM8K} & \textbf{MATH} & \textbf{HumanEval} \\
\midrule
\textbf{LoRA}                         & 16          &  89.33         &  88.58        &  87.06         &   86.73      &  73.16        &46.04       & 53.66      \\\hline
\midrule
\textbf{AQ-SGD 4}                       & 4           &  89.45	    &  88.24         &  85.16         &   86.65      &  71.80       & 41.92   & 55.49         \\ 
\textbf{AMAQ 4}                     & 4 ± 0.1     &  \textbf{89.63}       &  \textbf{88.41}         &  \textbf{86.42}	      &   \textbf{86.73}	     &  \textbf{72.10}       & \textbf{44.74}   & \textbf{56.71}      \\\hline
\midrule
\textbf{AQ-SGD 3}                       & 3           &  89.42        &  88.15         &  85.24	      &   87.06	     &  69.07        & 38.22       & 50.00   \\

\textbf{AMAQ 3}                     &  3 ± 0.1    &  \textbf{89.57}	    &  \textbf{88.67}         &  \textbf{86.42}         &   \textbf{87.14}     &  \textbf{72.33}        & \textbf{42.04}       & \textbf{51.22}       \\\hline

\bottomrule
\end{tabular}}
\caption{Qwen 2.5-7B evaluation using AMAQ for both input and output activations.}\label{QwenPerformance}
\end{table*}

\section{Results}

\subsection{AMAQ for Generation Tasks}
We evaluate the activation quantization methods using perplexity as the primary metric across LLaMA3-8B, Qwen2.5-14B, and Phi-3-Medium models as summarized in Table~\ref{ppl}. Our approach consistently outperforms AQ-SGD and other baselines, demonstrating notable improvements on generation tasks such as GSM8K, MATH, and Code-Alpaca. This is also evident in Figure~\ref{plot}, where the test loss for each task indicates that our approach, AMAQ, outperforms the 4-bit AQ-SGD quantization, although it slightly trails behind half-precision activation performance. For example, in tasks like GSM8K and MATH, the quantization level reaches 4 bits from 8 bits in roughly 500 steps. However, more challenging tasks with larger training datasets, such as Code-Alpaca, require approximately 1200 steps to achieve 4-bit quantization, using the same $\beta$ value. 

Furthermore, 
the Appendix shows that adapting AMAQ with 4-bit activations and 3-bit activation quantization in Qwen2.5 7B achieves a comparable best test loss to BF16, marking a significant improvement. 

\begin{figure}[!t]
    \centering
    \includegraphics[width=0.95\linewidth]{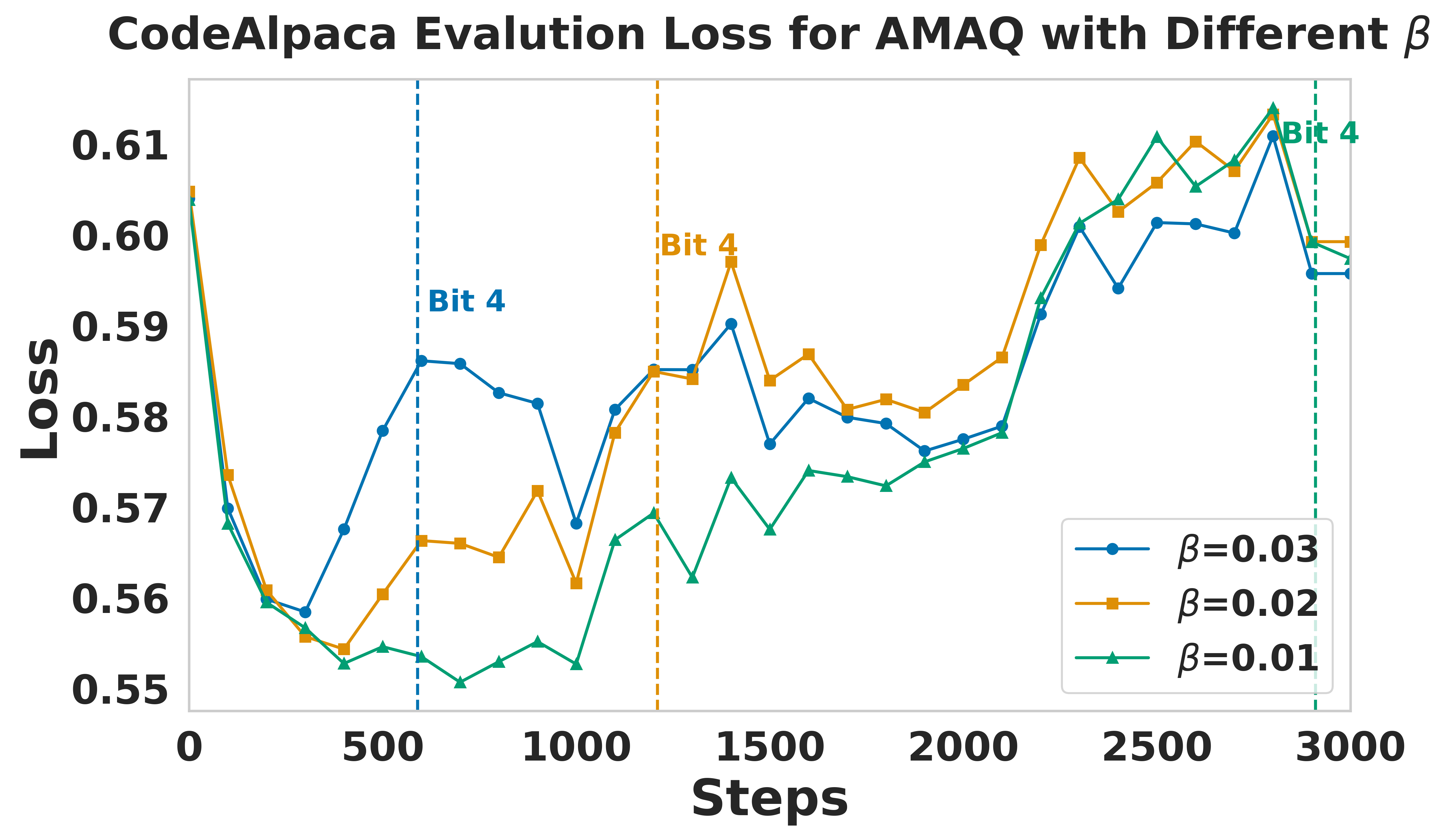}
     \caption{Performance of CodeAlpaca with different $\beta$} \label{beta}
     \vspace{-1em}
\end{figure}




We further evaluate Exact Match accuracy on GSM8K and MATH, as well as Pass@1 on HumanEval, following fine-tuning with both AQ-SGD quantization and our AMAQ approach in Table~\ref{LLaMA3Performance} and Table~\ref{QwenPerformance}. On LLaMA3-8B, our method achieves improvements of 1.06\% on GSM8K, 0.76\% on MATH, and 5.48\% on HumanEval. On Qwen2.5-8B, we observe a 3.8\% gain on MATH and a 1.22\% increase on HumanEval under the same 4-bit budget. Meanwhile, for 3-bit quantization, our method delivers consistent performance gains ranging from 1.78\% to 2.58\% across both LLaMA3-8B and Qwen2.5-7B models on average.



\begin{figure}[!ht]
    \centering
    \includegraphics[width=\linewidth]{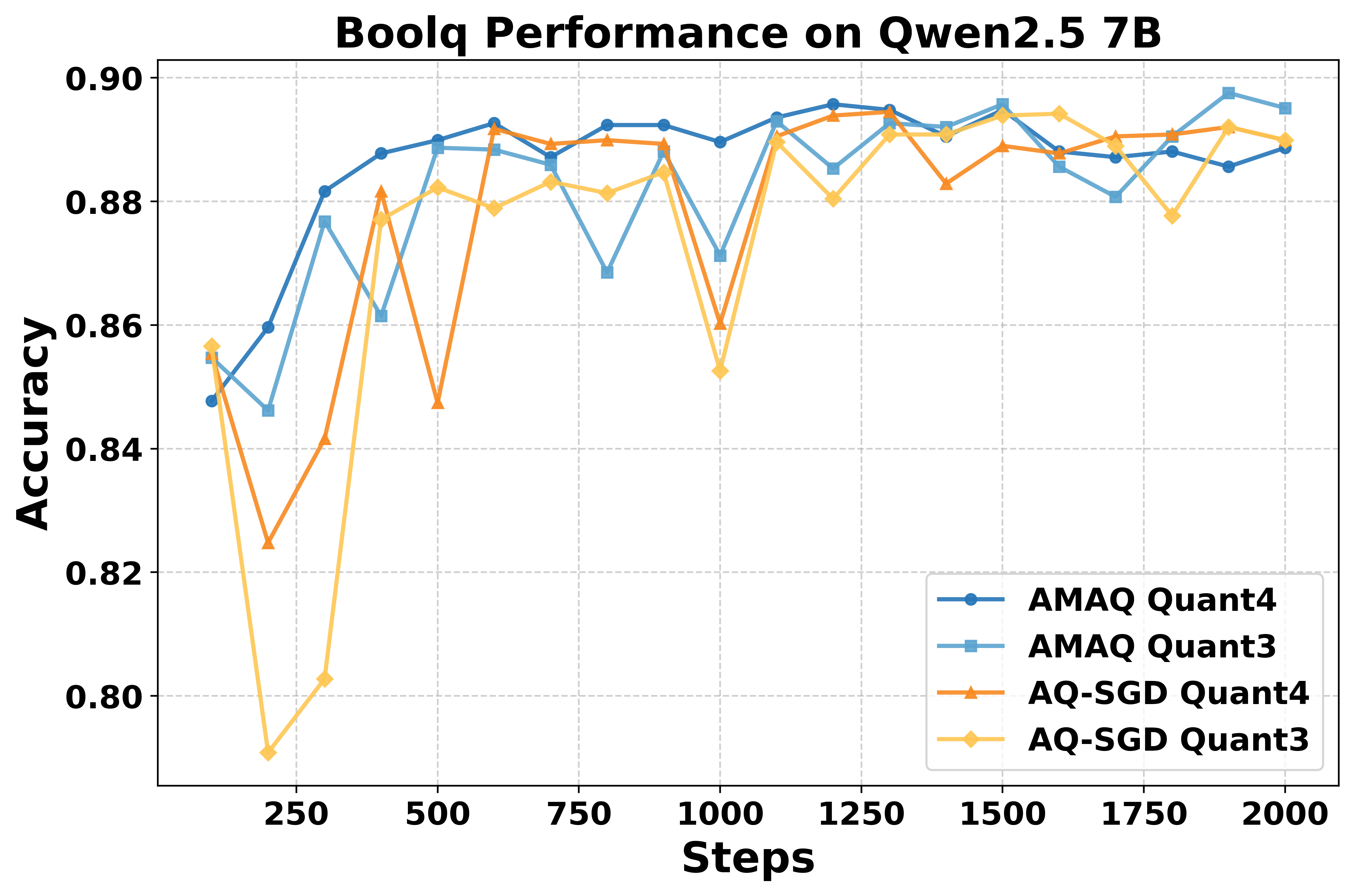}
     \caption{Performance and stability of Qwen2.5-7B
on BoolQ under input and output activation quantization.} \label{boolq}
\end{figure}

\begin{table*}[!ht]
\centering
\scalebox{0.85}{
\begin{tabular}{lllll}
\toprule
\multicolumn{1}{c|}{\textbf{LLaMA3-8B}}                            & \multicolumn{1}{l|}{\textbf{GSM8K}} & \multicolumn{1}{l|}{\textbf{Client Memory}} & \multicolumn{1}{l|}{\textbf{Communication Latency}} & \multicolumn{1}{l}{\textbf{Transmission Size} (/ batch)} \\ \hline
\multicolumn{5}{c}{\textbf{LoRA}}                                                                                                                                             \\ \hline
\multicolumn{1}{l|}{Client + Server BF16 }                & \multicolumn{1}{c|}{55.12}      & \multicolumn{1}{c|}{609 MB}          & \multicolumn{1}{c|}{1.18 X}            &     \multicolumn{1}{c}{27.3 MB}                         \\ 
\multicolumn{1}{l|}{Client + Server AQ-SGD 4}               & \multicolumn{1}{c|}{53.15}      & \multicolumn{1}{c|}{643 MB}          & \multicolumn{1}{c|}{1.00 X}            &     \multicolumn{1}{c}{6.8 MB}                          \\ 
\multicolumn{1}{l|}{Client + Server AMAQ 4 }               & \multicolumn{1}{c|}{54.28}      & \multicolumn{1}{c|}{681 MB}          & \multicolumn{1}{c|}{1.06 X}            &     \multicolumn{1}{c}{7.4 MB}                         \\ \hline
\multicolumn{5}{c}{\textbf{Activations + Gradient +  LoRA}}                                                                                                                                        \\ \hline
\multicolumn{1}{l|}{Client + Server BF16}                 & \multicolumn{1}{c|}{53.75}       & \multicolumn{1}{c|}{3.9 GB}          & \multicolumn{1}{c|}{1.63 X}      &   \multicolumn{1}{c}{226.5 MB}                          \\ 
\multicolumn{1}{l|}{Client + Server AQ-SGD 4}               & \multicolumn{1}{c|}{52.38}      & \multicolumn{1}{c|}{3.9 GB}          & \multicolumn{1}{c|}{1.00 X}       &   \multicolumn{1}{c}{56.4 MB}                       \\ 
\multicolumn{1}{l|}{Client + Server AMAQ 4}                & \multicolumn{1}{c|}{53.14}     & \multicolumn{1}{c|}{4.0 GB}          & \multicolumn{1}{c|}{1.14 X}       &   \multicolumn{1}{c}{61.4 MB}                         \\ 
\bottomrule
\end{tabular}}
\caption{Two-machine collaborative training in Figure~\ref{fig:use_case} (b) uses batch size 16 and LoRA rank 16, transmitting all Intermediate LoRA weights, activations, and gradients across networks using AMAQ quantization.} \label{distribute}
\end{table*}

\begin{table}[!ht]
\centering
\scalebox{0.7}{
\begin{tabular}{cc|c|c|c}
\toprule
\multicolumn{2}{c|}{\textbf{PPL} $\downarrow$} & \textbf{QAT-Group} & \textbf{AQ-SGD} & \textbf{AMAQ} \\ 
\midrule
\multirow{3}{*}{\textbf{LLaMA3 8B}}   & GSM8k       & 1.574       &1.576       &   \textbf{1.568}     \\ 
                                      & MATH        & 1.855       & 1.853       &   \textbf{1.837}     \\  
                                      & Code-Alpaca & \textbf{1.786}       &  1.790       &  1.791      \\ 
\midrule
\multirow{3}{*}{\textbf{Qwen2.5 7B}}  & GSM8k       & 1.590        & 1.593       &   \textbf{1.498}     \\ 
                                      & MATH        & 1.782        &   1.697         &   \textbf{1.668}     \\  
                                      & Code-Alpaca & 1.771        & 1.752       &  \textbf{1.701}      \\ 
\bottomrule
\end{tabular}}
\caption{4-bit Activation Compression with full finetuning using various quantization techniques. Activation compression is applied to the input and output activation only.} \label{full_finetune}
\end{table}

\subsection{AMAQ for Classification Tasks}
We investigate the performance of AMAQ on classification tasks by fine-tuning the LoRA module with bit budgets of 4 bits and 3 bits. Table~\ref{LLaMA3Performance} presents the results for the LLaMA3 8B model, while Table~\ref{QwenPerformance} shows the performance for the Qwen2.5 7B model. Overall, the findings indicate that the AMAQ consistently outperforms AQ-SGD across all evaluated tasks. In some instances, the performance improvement is marginal, around 0.2\%, but for other tasks the gains are more substantial, reaching up to 1.5\% for Winogrande and 3.7\% for the ARC Challenge. These differences appear to be highly dependent on the difficulty of the tasks. In the complex generation tasks, we observe a drop in performance when the quantization level drops from 4 to 3. However, in the simpler classification tasks, LLMs perform relatively well even at lower precisions.

\subsection{Server and Client Split Learning}




As shown in Table~\ref{distribute}, we evaluate the our framework under two deployment scenarios. In the first case, we implement intermediate LoRA, where only LoRA modules are transmitted over the network. In this setup, only BF16 PEFT modules are trained locally, requiring approximately 600MB of GPU memory and just 6.8MB of data transmitted per batch (including inputs and labels). When applying AMAQ, the per-batch transmitted size increases by only 0.5MB, with a minor 6\% increase in training time, but yields a 1.2\% accuracy gain on GSM8K in comparison with AQ-SGD. This is because in this setup, we clip at 4 bits, clipping to fewer bits or disabling clipping can further reduce communication and training time.


In the second case, we additionally deploy both the word embeds and the LM head locally. This configuration consumes approximately 4GB of GPU memory. Incorporating AMAQ in this setting increases per-batch transmission size by only 5MB compared to AQ-SGD quantization, with a 10\% training time overhead. Despite this, it achieves a 0.7\% accuracy improvement on GSM8K over AQ-SGD, and performs only 0.7\% below the full BF16 baseline.
Overall, our framework demonstrates an effective balance between performance and efficiency, making it practical for deployment on low-resource devices.

\subsection{Full Fine-Tuning vs. LoRA Adaptation}
We extend our evaluation by applying full finetuning to all layers rather than just LoRA modules. Table \ref{full_finetune} summarizes perplexity across three tasks for QAT-Group, AQ-SGD, and our AMAQ. On average, AMAQ reduces PPL by 2\% – 4\% relative to QAT-Group and by 1\% – 3\% relative to AQ-SGD, demonstrating its consistent advantage under full finetuning. While Table \ref{ppl} reports results when LoRA is tuned in the intermediate layers. 

Full finetuning with AMAQ outperforms the LoRA baseline. We find that convergence in Code-Alpaca is substantially slower than in other benchmarks (e.g., MATH), indicating the need for additional hyperparameter tuning, especially for AMAQ quantization, since we applied identical AMAQ settings to both full fine-tuning and LoRA.

\subsection{All-Layer Activation Quantization}
To fully assess the effectiveness of AMAQ, we explore the impact of adapting AMAQ across all transformer layers rather than the first and last layers. This setting is more challenging and introduces a lot of instability during training. To evaluate the robustness of AMAQ, we conduct our experiments using a larger model, Qwen2.5 14B.

As shown in Table~\ref{all_layer}, AMAQ consistently outperforms AQ-SGD under both 4-bit and 3-bit activation quantization. For instance, even in the BoolQ task, while AQ-SGD performs comparably to AMAQ at 3-bit precision, AMAQ shows significantly more stable training at both 4-bit and 3-bit levels, as illustrated in Figure~\ref{boolq_all}.

However, we find that applying AMAQ across all layers requires more careful and nuanced hyperparameter tuning. In particular, reducing the bit-width across all layers to the target bits level requires more training steps than tuning only the first and last layers. 
As the number of quantized activation layers increases, it becomes more challenging to reduce to an ultra-low bit-width effectively. It highlight the need for more effective hyperparameter tuning strategies when extending AMAQ to quantize all layers.



\section{Analysis and Discussion}

\subsection{Stability with Quantization Activation}




One key advantage of AMAQ is its improved training stability. As shown in Figure~\ref{boolq}, AMAQ outperforms other activation quantization methods in terms of stability. While AQ-SGD methods can occasionally achieve good results, their training behavior, particularly at 3-bit precision, is highly unstable, with significant fluctuations during the early stages of training. This highlights the importance of a stable initialization strategy. 

We also observe that instability grows as more layers are fully quantized. In the Qwen2.5-14B model with all layers activation quantization, AQ-SGD leads to noticeable instability and even divergence,
whereas AMAQ offers relatively stable training dynamics, even under low-bit, full-layer quantization settings.

\subsection{Bit-Width Adaptation Speed}
\label{ref:adaptation-speed}

We investigate the impact of different $\beta$ values on performance in Figure~\ref{beta} and find that $\beta$ significantly influences the speed of reaching the target bit level, although a faster speed up may come at the cost of some performance degradation. For instance, setting $\beta=0.01$ allowed the system to reach 4 bits in about 3000 steps, whereas $\beta=0.03$ achieves the same level in roughly 600 steps. But in general, performance with $\beta=0.01$ slightly outperforms that of $\beta=0.03$. The sensitivity to $\beta$ depends on the learning rate and specific task.

\subsection{Uneven Bit Allocation Across Layers}




We find that the input and output layers exhibit different sensitivities to quantization. On average, for generation tasks, the output layer requires approximately one bit higher width than the input layer. This indicates that the input data becomes relatively more obscured through quantization, while the output retains a higher level of detail. Consequently, this asymmetry in activation quantization preserves more granular information in the output for task performance.

\subsection{Sensitivity of Hyperparameters}\label{ref:hyperparameters}



In addition to tuning $\beta$, 
we also experiment with different learning rates for AMAQ. A higher learning rate and a higher value of $\beta$ can speed up convergence to lower-bit precision. However, it introduces the risk of training instability. Finding the optimal balance between these parameters remains an important area for future work. In practice, we devise a robust hyperparameter‐tuning strategy of beginning with a large learning rate and $\beta$ so the model reaches its target bit budget before overfitting, and then progressively decreasing these hyperparameters to enhance the stability and maximize performance.

\section{Conclusion}

In this work, we introduce Adaptive Mixed-bit Activation Quantization (AMAQ), a novel approach that dynamically adjusts activation bit-widths from high to low precision based on feature and layer importance. Designed for distributed collaborative training, AMAQ significantly reduces communication overhead while outperforming fixed-bit methods under equivalent transmission budgets. Our comprehensive evaluation demonstrates AMAQ's effectiveness across both LoRA and full fine-tuning paradigms, in single-machine and multi-machine environments, and for both input\&output-layer and All-Layer quantization settings. Results consistently show that AMAQ's adaptive bit allocation preserves model performance more effectively than existing quantization techniques. Future work will explore multi-party collaborative training framework and develop hyperparameter sensitivity analysis strategies.

\clearpage


\bibliography{anthology,custom}

\begin{thebibliography}{36}
\expandafter\ifx\csname natexlab\endcsname\relax\def\natexlab#1{#1}\fi

\bibitem[{Chaudhary(2023)}]{codealpaca}
Sahil Chaudhary. 2023.
\newblock Code alpaca: An instruction-following llama model for code generation.
\newblock \url{https://github.com/sahil280114/codealpaca}.

\bibitem[{Chen et~al.(2021)Chen, Tworek, Jun, Yuan, de~Oliveira~Pinto, Kaplan, Edwards, Burda, Joseph, Brockman, Ray, Puri, Krueger, Petrov, Khlaaf, Sastry, Mishkin, Chan, Gray, Ryder, Pavlov, Power, Kaiser, Bavarian, Winter, Tillet, Such, Cummings, Plappert, Chantzis, Barnes, Herbert-Voss, Guss, Nichol, Paino, Tezak, Tang, Babuschkin, Balaji, Jain, Saunders, Hesse, Carr, Leike, Achiam, Misra, Morikawa, Radford, Knight, Brundage, Murati, Mayer, Welinder, McGrew, Amodei, McCandlish, Sutskever, and Zaremba}]{chen2021codex}
Mark Chen, Jerry Tworek, Heewoo Jun, Qiming Yuan, Henrique~Ponde de~Oliveira~Pinto, Jared Kaplan, Harri Edwards, Yuri Burda, Nicholas Joseph, Greg Brockman, Alex Ray, Raul Puri, Gretchen Krueger, Michael Petrov, Heidy Khlaaf, Girish Sastry, Pamela Mishkin, Brooke Chan, Scott Gray, Nick Ryder, Mikhail Pavlov, Alethea Power, Lukasz Kaiser, Mohammad Bavarian, Clemens Winter, Philippe Tillet, Felipe~Petroski Such, Dave Cummings, Matthias Plappert, Fotios Chantzis, Elizabeth Barnes, Ariel Herbert-Voss, William~Hebgen Guss, Alex Nichol, Alex Paino, Nikolas Tezak, Jie Tang, Igor Babuschkin, Suchir Balaji, Shantanu Jain, William Saunders, Christopher Hesse, Andrew~N. Carr, Jan Leike, Josh Achiam, Vedant Misra, Evan Morikawa, Alec Radford, Matthew Knight, Miles Brundage, Mira Murati, Katie Mayer, Peter Welinder, Bob McGrew, Dario Amodei, Sam McCandlish, Ilya Sutskever, and Wojciech Zaremba. 2021.
\newblock \href {http://arxiv.org/abs/2107.03374} {Evaluating large language models trained on code}.

\bibitem[{Chen et~al.(2024)Chen, Xie, Li, and Shen}]{chen2024channel}
Zihan Chen, Bike Xie, Jundong Li, and Cong Shen. 2024.
\newblock Channel-wise mixed-precision quantization for large language models.
\newblock \emph{arXiv preprint arXiv:2410.13056}.

\bibitem[{Choi et~al.(2018)Choi, Wang, Venkataramani, Chuang, Srinivasan, and Gopalakrishnan}]{choi2018pactparameterizedclippingactivation}
Jungwook Choi, Zhuo Wang, Swagath Venkataramani, Pierce I-Jen Chuang, Vijayalakshmi Srinivasan, and Kailash Gopalakrishnan. 2018.
\newblock \href {http://arxiv.org/abs/1805.06085} {Pact: Parameterized clipping activation for quantized neural networks}.

\bibitem[{Clark et~al.(2019)Clark, Lee, Chang, Kwiatkowski, Collins, and Toutanova}]{clark2019boolqexploringsurprisingdifficulty}
Christopher Clark, Kenton Lee, Ming-Wei Chang, Tom Kwiatkowski, Michael Collins, and Kristina Toutanova. 2019.
\newblock \href {http://arxiv.org/abs/1905.10044} {Boolq: Exploring the surprising difficulty of natural yes/no questions}.

\bibitem[{Clark et~al.(2018)Clark, Cowhey, Etzioni, Khot, Sabharwal, Schoenick, and Tafjord}]{allenai:arc}
Peter Clark, Isaac Cowhey, Oren Etzioni, Tushar Khot, Ashish Sabharwal, Carissa Schoenick, and Oyvind Tafjord. 2018.
\newblock Think you have solved question answering? try arc, the ai2 reasoning challenge.
\newblock \emph{arXiv:1803.05457v1}.

\bibitem[{Cobbe et~al.(2021)Cobbe, Kosaraju, Bavarian, Chen, Jun, Kaiser, Plappert, Tworek, Hilton, Nakano, Hesse, and Schulman}]{cobbe2021trainingverifierssolvemath}
Karl Cobbe, Vineet Kosaraju, Mohammad Bavarian, Mark Chen, Heewoo Jun, Lukasz Kaiser, Matthias Plappert, Jerry Tworek, Jacob Hilton, Reiichiro Nakano, Christopher Hesse, and John Schulman. 2021.
\newblock \href {http://arxiv.org/abs/2110.14168} {Training verifiers to solve math word problems}.

\bibitem[{DeepSeek-AI et~al.(2025)DeepSeek-AI, Liu, Feng, Xue, Wang, Wu, Lu, Zhao, Deng, Zhang, Ruan, Dai, Guo, Yang, Chen, Ji, Li, Lin, Dai, Luo, Hao, Chen, Li, Zhang, Bao, Xu, Wang, Zhang, Ding, Xin, Gao, Li, Qu, Cai, Liang, Guo, Ni, Li, Wang, Chen, Chen, Yuan, Qiu, Li, Song, Dong, Hu, Gao, Guan, Huang, Yu, Wang, Zhang, Xu, Xia, Zhao, Wang, Zhang, Li, Wang, Zhang, Zhang, Tang, Li, Tian, Huang, Wang, Zhang, Wang, Zhu, Chen, Du, Chen, Jin, Ge, Zhang, Pan, Wang, Xu, Zhang, Chen, Li, Lu, Zhou, Chen, Wu, Ye, Ye, Ma, Wang, Zhou, Yu, Zhou, Pan, Wang, Yun, Pei, Sun, Xiao, Zeng, Zhao, An, Liu, Liang, Gao, Yu, Zhang, Li, Jin, Wang, Bi, Liu, Wang, Shen, Chen, Zhang, Chen, Nie, Sun, Wang, Cheng, Liu, Xie, Liu, Yu, Song, Shan, Zhou, Yang, Li, Su, Lin, Li, Wang, Wei, Zhu, Zhang, Xu, Xu, Huang, Li, Zhao, Sun, Li, Wang, Yu, Zheng, Zhang, Shi, Xiong, He, Tang, Piao, Wang, Tan, Ma, Liu, Guo, Wu, Ou, Zhu, Wang, Gong, Zou, He, Zha, Xiong, Ma, Yan, Luo, You, Liu, Zhou, Wu, Ren, Ren, Sha, Fu, Xu, Huang, Zhang, Xie, Zhang, Hao,
  Gou, Ma, Yan, Shao, Xu, Wu, Zhang, Li, Gu, Zhu, Liu, Li, Xie, Song, Gao, and Pan}]{deepseekai2025deepseekv3technicalreport}
DeepSeek-AI, Aixin Liu, Bei Feng, Bing Xue, Bingxuan Wang, Bochao Wu, Chengda Lu, Chenggang Zhao, Chengqi Deng, Chenyu Zhang, Chong Ruan, Damai Dai, Daya Guo, Dejian Yang, Deli Chen, Dongjie Ji, Erhang Li, Fangyun Lin, Fucong Dai, Fuli Luo, Guangbo Hao, Guanting Chen, Guowei Li, H.~Zhang, Han Bao, Hanwei Xu, Haocheng Wang, Haowei Zhang, Honghui Ding, Huajian Xin, Huazuo Gao, Hui Li, Hui Qu, J.~L. Cai, Jian Liang, Jianzhong Guo, Jiaqi Ni, Jiashi Li, Jiawei Wang, Jin Chen, Jingchang Chen, Jingyang Yuan, Junjie Qiu, Junlong Li, Junxiao Song, Kai Dong, Kai Hu, Kaige Gao, Kang Guan, Kexin Huang, Kuai Yu, Lean Wang, Lecong Zhang, Lei Xu, Leyi Xia, Liang Zhao, Litong Wang, Liyue Zhang, Meng Li, Miaojun Wang, Mingchuan Zhang, Minghua Zhang, Minghui Tang, Mingming Li, Ning Tian, Panpan Huang, Peiyi Wang, Peng Zhang, Qiancheng Wang, Qihao Zhu, Qinyu Chen, Qiushi Du, R.~J. Chen, R.~L. Jin, Ruiqi Ge, Ruisong Zhang, Ruizhe Pan, Runji Wang, Runxin Xu, Ruoyu Zhang, Ruyi Chen, S.~S. Li, Shanghao Lu, Shangyan Zhou, Shanhuang
  Chen, Shaoqing Wu, Shengfeng Ye, Shengfeng Ye, Shirong Ma, Shiyu Wang, Shuang Zhou, Shuiping Yu, Shunfeng Zhou, Shuting Pan, T.~Wang, Tao Yun, Tian Pei, Tianyu Sun, W.~L. Xiao, Wangding Zeng, Wanjia Zhao, Wei An, Wen Liu, Wenfeng Liang, Wenjun Gao, Wenqin Yu, Wentao Zhang, X.~Q. Li, Xiangyue Jin, Xianzu Wang, Xiao Bi, Xiaodong Liu, Xiaohan Wang, Xiaojin Shen, Xiaokang Chen, Xiaokang Zhang, Xiaosha Chen, Xiaotao Nie, Xiaowen Sun, Xiaoxiang Wang, Xin Cheng, Xin Liu, Xin Xie, Xingchao Liu, Xingkai Yu, Xinnan Song, Xinxia Shan, Xinyi Zhou, Xinyu Yang, Xinyuan Li, Xuecheng Su, Xuheng Lin, Y.~K. Li, Y.~Q. Wang, Y.~X. Wei, Y.~X. Zhu, Yang Zhang, Yanhong Xu, Yanhong Xu, Yanping Huang, Yao Li, Yao Zhao, Yaofeng Sun, Yaohui Li, Yaohui Wang, Yi~Yu, Yi~Zheng, Yichao Zhang, Yifan Shi, Yiliang Xiong, Ying He, Ying Tang, Yishi Piao, Yisong Wang, Yixuan Tan, Yiyang Ma, Yiyuan Liu, Yongqiang Guo, Yu~Wu, Yuan Ou, Yuchen Zhu, Yuduan Wang, Yue Gong, Yuheng Zou, Yujia He, Yukun Zha, Yunfan Xiong, Yunxian Ma, Yuting Yan, Yuxiang
  Luo, Yuxiang You, Yuxuan Liu, Yuyang Zhou, Z.~F. Wu, Z.~Z. Ren, Zehui Ren, Zhangli Sha, Zhe Fu, Zhean Xu, Zhen Huang, Zhen Zhang, Zhenda Xie, Zhengyan Zhang, Zhewen Hao, Zhibin Gou, Zhicheng Ma, Zhigang Yan, Zhihong Shao, Zhipeng Xu, Zhiyu Wu, Zhongyu Zhang, Zhuoshu Li, Zihui Gu, Zijia Zhu, Zijun Liu, Zilin Li, Ziwei Xie, Ziyang Song, Ziyi Gao, and Zizheng Pan. 2025.
\newblock \href {http://arxiv.org/abs/2412.19437} {Deepseek-v3 technical report}.

\bibitem[{Evci et~al.(2021)Evci, Gale, Menick, Castro, and Elsen}]{rigL}
Utku Evci, Trevor Gale, Jacob Menick, Pablo~Samuel Castro, and Erich Elsen. 2021.
\newblock \href {http://arxiv.org/abs/1911.11134} {Rigging the lottery: Making all tickets winners}.

\bibitem[{Gao and Zhang(2024)}]{gao2024dloradistributedparameterefficientfinetuning}
Chao Gao and Sai~Qian Zhang. 2024.
\newblock \href {http://arxiv.org/abs/2404.05182} {Dlora: Distributed parameter-efficient fine-tuning solution for large language model}.

\bibitem[{Hendrycks et~al.(2021)Hendrycks, Burns, Kadavath, Arora, Basart, Tang, Song, and Steinhardt}]{hendrycksmath2021}
Dan Hendrycks, Collin Burns, Saurav Kadavath, Akul Arora, Steven Basart, Eric Tang, Dawn Song, and Jacob Steinhardt. 2021.
\newblock Measuring mathematical problem solving with the math dataset.
\newblock \emph{NeurIPS}.

\bibitem[{Houlsby et~al.(2019)Houlsby, Giurgiu, Jastrzebski, Morrone, de~Laroussilhe, Gesmundo, Attariyan, and Gelly}]{adaptor-modules}
Neil Houlsby, Andrei Giurgiu, Stanislaw Jastrzebski, Bruna Morrone, Quentin de~Laroussilhe, Andrea Gesmundo, Mona Attariyan, and Sylvain Gelly. 2019.
\newblock \href {http://arxiv.org/abs/1902.00751} {Parameter-efficient transfer learning for nlp}.

\bibitem[{Hu et~al.(2021)Hu, Shen, Wallis, Allen-Zhu, Li, Wang, Wang, and Chen}]{hu2021loralowrankadaptationlarge}
Edward~J. Hu, Yelong Shen, Phillip Wallis, Zeyuan Allen-Zhu, Yuanzhi Li, Shean Wang, Lu~Wang, and Weizhu Chen. 2021.
\newblock \href {http://arxiv.org/abs/2106.09685} {Lora: Low-rank adaptation of large language models}.

\bibitem[{Hubara et~al.(2021)Hubara, Nahshan, Hanani, Banner, and Soudry}]{Min-Max-Quantization}
Itay Hubara, Yury Nahshan, Yair Hanani, Ron Banner, and Daniel Soudry. 2021.
\newblock \href {https://proceedings.mlr.press/v139/hubara21a.html} {Accurate post training quantization with small calibration sets}.
\newblock In \emph{Proceedings of the 38th International Conference on Machine Learning}, volume 139 of \emph{Proceedings of Machine Learning Research}, pages 4466--4475. PMLR.

\bibitem[{Jain et~al.(2020)Jain, Gural, Wu, and Dick}]{KL-Divergence}
Sambhav~R. Jain, Albert Gural, Michael Wu, and Chris~H. Dick. 2020.
\newblock \href {http://arxiv.org/abs/1903.08066} {Trained quantization thresholds for accurate and efficient fixed-point inference of deep neural networks}.

\bibitem[{Li and Liang(2021{\natexlab{a}})}]{li2021prefixtuningoptimizingcontinuousprompts}
Xiang~Lisa Li and Percy Liang. 2021{\natexlab{a}}.
\newblock \href {http://arxiv.org/abs/2101.00190} {Prefix-tuning: Optimizing continuous prompts for generation}.

\bibitem[{Li and Liang(2021{\natexlab{b}})}]{prefix-tuning}
Xiang~Lisa Li and Percy Liang. 2021{\natexlab{b}}.
\newblock \href {http://arxiv.org/abs/2101.00190} {Prefix-tuning: Optimizing continuous prompts for generation}.

\bibitem[{Li et~al.(2022)Li, Tramer, Liang, and Hashimoto}]{li2022large}
Xuechen Li, Florian Tramer, Percy Liang, and Tatsunori Hashimoto. 2022.
\newblock \href {https://openreview.net/forum?id=bVuP3ltATMz} {Large language models can be strong differentially private learners}.
\newblock In \emph{Proceedings of the International Conference on Learning Representations (ICLR)}.

\bibitem[{Li et~al.(2023)Li, Tan, and Liu}]{li2023privacy}
Yansong Li, Zhixing Tan, and Yang Liu. 2023.
\newblock \href {https://api.semanticscholar.org/CorpusID:258588141} {Privacy-preserving prompt tuning for large language model services}.
\newblock \emph{arXiv preprint arXiv:2305.06212}.

\bibitem[{Lin et~al.(2024{\natexlab{a}})Lin, Tang, Tang, Yang, Chen, Wang, Xiao, Dang, Gan, and Han}]{lin2024awqactivationawareweightquantization}
Ji~Lin, Jiaming Tang, Haotian Tang, Shang Yang, Wei-Ming Chen, Wei-Chen Wang, Guangxuan Xiao, Xingyu Dang, Chuang Gan, and Song Han. 2024{\natexlab{a}}.
\newblock \href {http://arxiv.org/abs/2306.00978} {Awq: Activation-aware weight quantization for llm compression and acceleration}.

\bibitem[{Lin et~al.(2022)Lin, Li, Cheng, Kuo, Lu, and Tang}]{lin2022lglsqlearnedgradientlinear}
Shih-Ting Lin, Zhaofang Li, Yu-Hsiang Cheng, Hao-Wen Kuo, Chih-Cheng Lu, and Kea-Tiong Tang. 2022.
\newblock \href {http://arxiv.org/abs/2202.09009} {Lg-lsq: Learned gradient linear symmetric quantization}.

\bibitem[{Lin et~al.(2024{\natexlab{b}})Lin, Hu, Zhang, Chen, Fang, Chen, Li, Vepakomma, and Gao}]{lin2024splitlorasplitparameterefficientfinetuning}
Zheng Lin, Xuanjie Hu, Yuxin Zhang, Zhe Chen, Zihan Fang, Xianhao Chen, Ang Li, Praneeth Vepakomma, and Yue Gao. 2024{\natexlab{b}}.
\newblock \href {http://arxiv.org/abs/2407.00952} {Splitlora: A split parameter-efficient fine-tuning framework for large language models}.

\bibitem[{Liu et~al.(2022)Liu, Ji, Fu, Tam, Du, Yang, and Tang}]{liu2022ptuningv2prompttuning}
Xiao Liu, Kaixuan Ji, Yicheng Fu, Weng~Lam Tam, Zhengxiao Du, Zhilin Yang, and Jie Tang. 2022.
\newblock \href {http://arxiv.org/abs/2110.07602} {P-tuning v2: Prompt tuning can be comparable to fine-tuning universally across scales and tasks}.

\bibitem[{Nagel et~al.(2019)Nagel, van Baalen, Blankevoort, and Welling}]{Bias-Correction}
Markus Nagel, Mart van Baalen, Tijmen Blankevoort, and Max Welling. 2019.
\newblock \href {http://arxiv.org/abs/1906.04721} {Data-free quantization through weight equalization and bias correction}.

\bibitem[{Sakaguchi et~al.(2019)Sakaguchi, Bras, Bhagavatula, and Choi}]{sakaguchi2019winograndeadversarialwinogradschema}
Keisuke Sakaguchi, Ronan~Le Bras, Chandra Bhagavatula, and Yejin Choi. 2019.
\newblock \href {http://arxiv.org/abs/1907.10641} {Winogrande: An adversarial winograd schema challenge at scale}.

\bibitem[{Sun et~al.(2019)Sun, Ren, Ma, and Wang}]{meProp}
Xu~Sun, Xuancheng Ren, Shuming Ma, and Houfeng Wang. 2019.
\newblock \href {http://arxiv.org/abs/1706.06197} {meprop: Sparsified back propagation for accelerated deep learning with reduced overfitting}.

\bibitem[{Talmor et~al.(2019)Talmor, Herzig, Lourie, and Berant}]{talmor-etal-2019-commonsenseqa}
Alon Talmor, Jonathan Herzig, Nicholas Lourie, and Jonathan Berant. 2019.
\newblock \href {https://doi.org/10.18653/v1/N19-1421} {{C}ommonsense{QA}: A question answering challenge targeting commonsense knowledge}.
\newblock In \emph{Proceedings of the 2019 Conference of the North {A}merican Chapter of the Association for Computational Linguistics: Human Language Technologies, Volume 1 (Long and Short Papers)}, pages 4149--4158, Minneapolis, Minnesota. Association for Computational Linguistics.

\bibitem[{Tay et~al.(2020)Tay, Bahri, Yang, Metzler, and Juan}]{sparsesinkhornattention}
Yi~Tay, Dara Bahri, Liu Yang, Donald Metzler, and Da-Cheng Juan. 2020.
\newblock \href {http://arxiv.org/abs/2002.11296} {Sparse sinkhorn attention}.

\bibitem[{Thapa et~al.(2022)Thapa, Arachchige, Camtepe, and Sun}]{thapa2022splitfed}
Chandra Thapa, Pathum Chamikara~Mahawaga Arachchige, Seyit Camtepe, and Lichao Sun. 2022.
\newblock Splitfed: When federated learning meets split learning.
\newblock In \emph{Proceedings of the AAAI conference on artificial intelligence}, volume~36, pages 8485--8493.

\bibitem[{Wang et~al.(2023)Wang, Yuan, Rimanic, He, Dao, Chen, Re, and Zhang}]{wang2023finetuninglanguagemodelsslow}
Jue Wang, Binhang Yuan, Luka Rimanic, Yongjun He, Tri Dao, Beidi Chen, Christopher Re, and Ce~Zhang. 2023.
\newblock \href {http://arxiv.org/abs/2206.01299} {Fine-tuning language models over slow networks using activation compression with guarantees}.

\bibitem[{Wang et~al.(2025)Wang, Gong, Liu, Zhao, Yang, Guo, Zha, and Cheng}]{wang2025optimizinglargelanguagemodel}
Ruizhe Wang, Yeyun Gong, Xiao Liu, Guoshuai Zhao, Ziyue Yang, Baining Guo, Zhengjun Zha, and Peng Cheng. 2025.
\newblock \href {http://arxiv.org/abs/2501.17116} {Optimizing large language model training using fp4 quantization}.

\bibitem[{Wu et~al.(2023)Wu, Zheng, Liu, and Zheng}]{wu2023estimatormeetsequilibriumperspective}
Xiao-Ming Wu, Dian Zheng, Zuhao Liu, and Wei-Shi Zheng. 2023.
\newblock \href {http://arxiv.org/abs/2308.06689} {Estimator meets equilibrium perspective: A rectified straight through estimator for binary neural networks training}.

\bibitem[{Xiao et~al.(2024)Xiao, Lin, Seznec, Wu, Demouth, and Han}]{xiao2024smoothquantaccurateefficientposttraining}
Guangxuan Xiao, Ji~Lin, Mickael Seznec, Hao Wu, Julien Demouth, and Song Han. 2024.
\newblock \href {http://arxiv.org/abs/2211.10438} {Smoothquant: Accurate and efficient post-training quantization for large language models}.

\bibitem[{Yang et~al.(2024)Yang, Tang, Yu, and Lv}]{yang2024gwq}
Jiaming Yang, Chenwei Tang, Caiyang Yu, and Jiancheng Lv. 2024.
\newblock Gwq: Group-wise quantization framework for neural networks.
\newblock In \emph{Asian Conference on Machine Learning}, pages 1526--1541. PMLR.

\bibitem[{Yu et~al.(2022)Yu, Naik, Backurs, Gopi, Inan, Kamath, Kulkarni, Lee, Manoel, Wutschitz, Yekhanin, and Zhang}]{yu2022DP}
Da~Yu, Saurabh Naik, Arturs Backurs, Sivakanth Gopi, Huseyin~A. Inan, Gautam Kamath, Janardhan Kulkarni, Yin~Tat Lee, Andre Manoel, Lukas Wutschitz, Sergey Yekhanin, and Huishuai Zhang. 2022.
\newblock \href {https://openreview.net/forum?id=Q42f0dfjECO} {Differentially private fine-tuning of language models}.
\newblock In \emph{Proceedings of the International Conference on Learning Representations (ICLR)}.

\bibitem[{Zhou et~al.(2018)Zhou, Wu, Ni, Zhou, Wen, and Zou}]{zhou2018dorefanettraininglowbitwidth}
Shuchang Zhou, Yuxin Wu, Zekun Ni, Xinyu Zhou, He~Wen, and Yuheng Zou. 2018.
\newblock \href {http://arxiv.org/abs/1606.06160} {Dorefa-net: Training low bitwidth convolutional neural networks with low bitwidth gradients}.

\end{thebibliography}
\bibliographystyle{acl_natbib}

\appendix

\clearpage

\section{Appendix} \label{sec:appendix}

\subsection{Hyperparameters}

\begin{table*}[!ht]

\centering
\begin{tabular}{lcccccccc}
\toprule\cline{1-9}
Dataset    & Batch size & LR   & Steps & Source Length & Target Length & $\beta$ & Bits LR   & $\alpha$ \\ \cline{1-9}
MATH       & 16         & 1e-4 & 5000  & 512           & 512           & 0.02    & 1e-2      & $\alpha$ \\
CodeAlpaca & 16         & 1e-4 & 5000  & 512           & 1024          & 0.03    & 1e-2      & $\alpha$ \\
GSM8K      & 16         & 1e-4 & 5000  & 512           & 512           & 0.02    & 1e-2      & $\alpha$ \\ \cline{1-9}
\bottomrule
\end{tabular}
\caption{Hyperparameters used for generation tasks across all the models.} \label{gen_hp}
\end{table*}

\begin{table*}[!ht]
\centering
\scalebox{0.92}{
\begin{tabular}{lcccccccc}
\toprule \cline{1-9}
Model         & Dataset                & Batch size & LR   & Steps & Source Length            & $\beta$ (Q4 / Q3)         & Bits LR   & $\alpha$ \\ \cline{1-9}
LLaMA3 8B         & CommonSense        & 16         & 5e-5 & 3000  & 1024                      & 0.1 / 0.2        & 1e-2      & $\alpha$ \\
                   & Winogrande        & 16         & 5e-5 & 3000  & 1024                      & 0.02 / 0.02      & 1e-2      & $\alpha$ \\
                   & Boolq             & 16         & 5e-5 & 3000  & 1024                      & 0.2 / 0.2        & 1e-2      & $\alpha$ \\ 
                   & ARC-C             & 16         & 5e-5 & 3000  & 1024                      & 0.02 / 0.02       & 1e-2      & $\alpha$ \\ \midrule
Qwen2.5 7B         & CommonSense       & 16         & 5e-5 & 3000  & 1024                      & 0.2 / 0.05       & 1e-2        & $\alpha$ \\
                   & Winogrande        & 16         & 5e-5 & 3000  & 1024                      & 0.2 / 0.05       & 1e-2        & $\alpha$ \\
                   & Boolq             & 16         & 5e-5 & 3000  & 1024                      & 0.2 / 0.05       & 1e-2        & $\alpha$ \\ 
                   & ARC-C             & 16         & 5e-5 & 3000  & 1024                      & 0.2 / 0.02        & 1e-2         & $\alpha$ \\  
\cline{1-9}\bottomrule

\end{tabular}}
\caption{Hyperparameters used for classification tasks with LLaMA3 8B and Qwen2.5 7B.} \label{class_hp}
\end{table*}

Table~\ref{gen_hp}  and Table~\ref{class_hp} summarize the hyperparameters for our classification and generation experiments, respectively.
For the generation tasks, we apply LoRA with rank 64 and $\alpha$ = 16 (unless otherwise noted in the comparison experiments). In the classification tasks, we use LoRA with rank 16 and $\alpha$ = 16.
The classification hyperparameters $\beta$ vary with the chosen quantization level. In the 4-bit experiments, we begin with an 8-bit initial width and progressively quantize down to 4 bits. Likewise, in the 3-bit experiments, we start from a 6-bit initial width and adaptively reduce precision to 3 bits. 

One effective hyperparameter‐tuning strategy is to start with a high learning rate of $Q$ and $\beta$, allowing the model to hit its target bit budget before overfitting and then adaptively reduce both the learning rate of $Q$ and $\beta$ to improve stability and maximize performance. 
For both GSM8K and Math, we apply Top P 0.9 and temperature 0.6. For HumanEval, Top P 0.9 and temperature 0.2 for it.



\subsection{Intermediate Full Finetuning}
We extend our evaluation by applying full finetuning to the intermediate layers rather than intermediate PEFT modules. Table \ref{full_finetune} summarizes perplexity across three tasks for QAT-Group, AQ-SGD, and our AMAQ. On average, AMAQ reduces PPL by 2\% – 4\% relative to QAT-Group and by 1\% – 3\% relative to AQ-SGD, demonstrating its consistent advantage under full finetuning. Table \ref{ppl} reports results when LoRA is tuned in the intermediate layers. Full finetuning with AMAQ outperforms the LoRA baseline. We find that convergence in Code-Alpaca is substantially slower than in other benchmarks (e.g., MATH), indicating the need for additional hyperparameter tuning, especially for AMAQ quantization, since we applied identical AMAQ settings to both full fine-tuning and LoRA.

\subsection{Intermediate Prefix Tuning}
For classification tasks, we use 30 virtual tokens in prefix tuning. To improve efficiency, given the small prefix size, we avoid applying AMAQ separately to the past key and value tensors at each layer. Instead, we share two AMAQ modules across all layers---one for keys and one for values.
 Our experiments in Table~\ref{prefixPerformance} show that AMAQ significantly outperforms AQ-SGD across various classification tasks, particularly in the more challenging 3-bit activation quantization setting. While AQ-SGD often fails to maintain performance at this lower bit-width, AMAQ remains stable and yields strong results.
Given that the past key values from prefix tuning are effectively a variant of the KV cache used during inference, AMAQ suggests a promising direction for future research: adapting AMAQ for efficient and accurate KV cache quantization.

\begin{table*}[!ht]
\centering
\scalebox{0.9}{
\begin{tabular}{cc|c|c|c}
\toprule
\multicolumn{2}{c|}{\textbf{PPL} $\downarrow$} & \textbf{QAT-Group} & \textbf{AQ-SGD} & \textbf{AMAQ} \\ 
\midrule
\multirow{3}{*}{\textbf{LLaMA3 8B}}   & GSM8k       & 1.574       &1.576       &   \textbf{1.568}     \\ 
                                      & MATH        & 1.855       & 1.853       &   \textbf{1.837}     \\  
                                      & Code-Alpaca & \textbf{1.786}       &  1.790       &  1.791      \\ 
\midrule
\multirow{3}{*}{\textbf{Qwen2.5 7B}}  & GSM8k       & 1.590        & 1.593       &   \textbf{1.498}     \\ 
                                      & MATH        & 1.782        &   1.697         &   \textbf{1.668}     \\  
                                      & Code-Alpaca & 1.771        & 1.752       &  \textbf{1.701}      \\ 
\bottomrule
\end{tabular}}
\caption{4-bit Activation Compression with full finetuning using various quantization techniques (listed in ~\S~\ref{subsec:baselines}). Activation compression is applied to the input and output activation only.} \label{full_finetune}
\end{table*} 


\begin{table*}[!ht]
\centering
\scalebox{0.8}{
\begin{tabular}{cc|c|c|c|c}
\toprule
 \textbf{Qwen2.5-7B}                 & \textbf{Bits}  & \textbf{Boolq}       & \textbf{ARC-C} &  \textbf{Winogrande} & \textbf{CommonSenseQA} \\
\midrule
\textbf{LoRA}                         & 16          &  89.14                &  87.46          &  82.39               &   84.68                                   \\\hline
\midrule
\textbf{AQ-SGD 4}                & 4           &  \textbf{88.83}	            &  86.43          &  78.53              &   84.60                                       \\ 
\textbf{AMAQ 4}                     & 4 ± 0.1     &  88.41       &  \textbf{86.86}         &  \textbf{80.11}	       & \textbf{84.93}      \\\hline
\midrule
\textbf{AQ-SGD 3}                & 3           &  81.98       & 81.37        & 59.11	      &   78.78                          \\ 

\textbf{AMAQ 3}                     &  3 ± 0.1    &  \textbf{85.35}	    &  \textbf{87.63}         &  \textbf{72.84}        & \textbf{83.53}       \\\hline

\bottomrule
\end{tabular}}
\caption{Qwen 2.5 7B Prefix Tuning with 30 virtual tokens. Evaluation using AMAQ for both input and output activations.}\label{prefixPerformance}
\end{table*}

\subsection{All Layer Activation Quantization}
To fully assess the effectiveness of AMAQ, we explore the impact of adapting AMAQ across all transformer layers rather than the first and last layers. This setting is more challenging and introduces greater instability during training. To address this issue, we conduct our experiments using a larger model --- Qwen2.5 14B.

\begin{figure}[!ht]
    \centering
    \includegraphics[width=\linewidth]{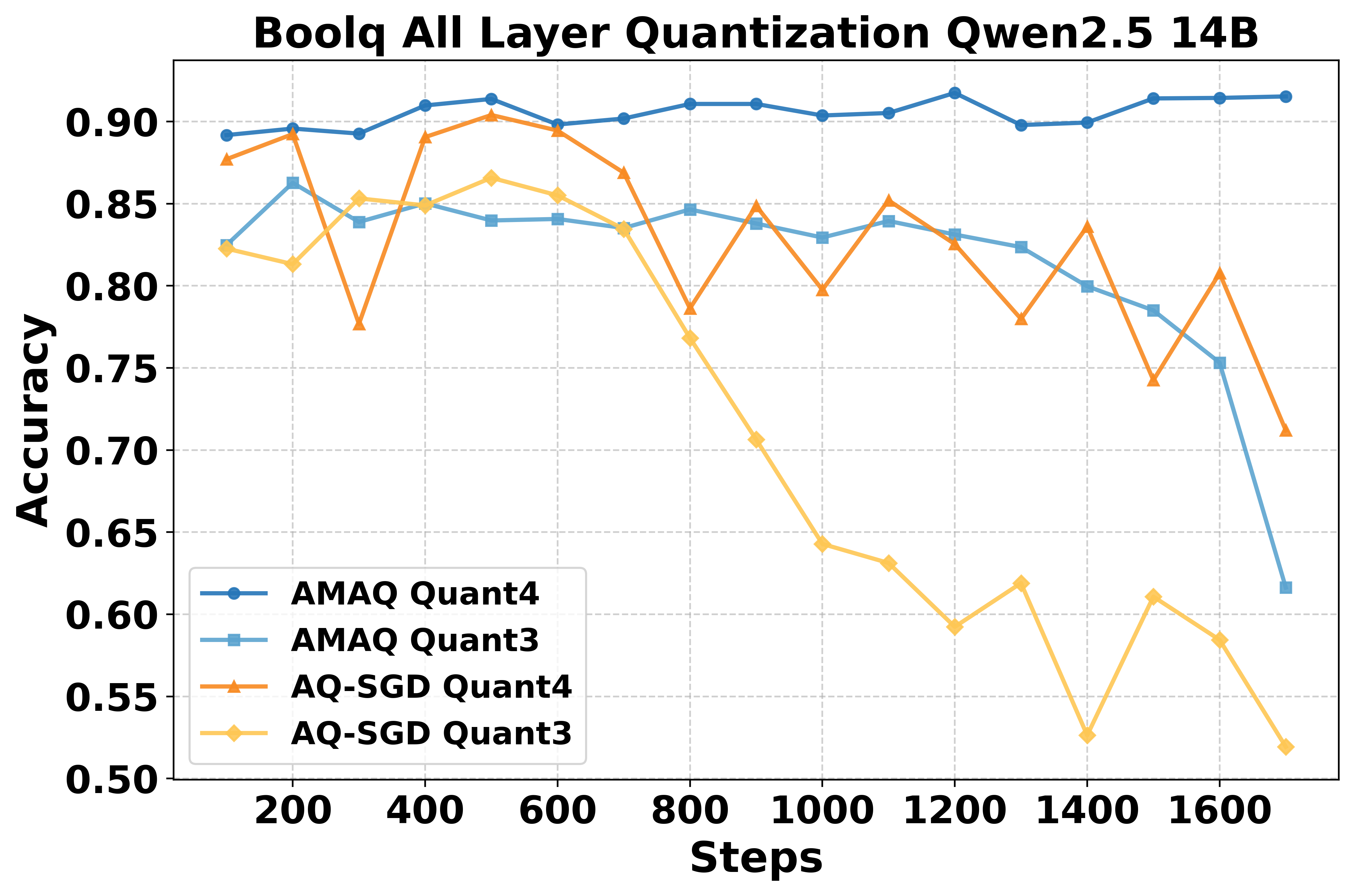}
     \caption{Performance and stability of Qwen2.5-14B on BoolQ under full-layer activation quantization.} \label{boolq_all}
\end{figure}

As shown in Table~\ref{all_layer}, AMAQ consistently outperforms AQ-SGD under both 4-bit and 3-bit activation quantization. For instance, even in the BoolQ task, while AQ-SGD performs comparably to AMAQ at 3-bit precision, AMAQ shows significantly more stable training at both 4-bit and 3-bit levels, as illustrated in Figure~\ref{boolq_all}.

However, we find that applying AMAQ across all layers requires more careful and nuanced hyperparameter tuning.
As the number of quantized layers increases, it becomes more challenging to reduce the bit-width effectively. \\
To address this, we observe that both the learning rate of $Q$ and the $\beta$ parameter need to be set higher to ensure the convergence. Moreover, reducing from 4 bits to 3 bits instead of starting from 6 bits helps fast convergence but might drop performance a bit.
These findings highlight the need for more effective hyperparameter tuning strategies when extending AMAQ to quantize all layers.

Our results can also contribute to ongoing efforts in FP4 Mixed Precision Training \cite{wang2025optimizinglargelanguagemodel}, where low-bit activation compression is critical for fitting large models and can help stable and high-performing Mixed Precision Training.

\begin{table*}[!ht]
\centering
\scalebox{0.8}{
\begin{tabular}{cc|c|c|c|c}
\toprule
 \textbf{Qwen2.5-14B}                 & \textbf{Bits}  & \textbf{Boolq}       & \textbf{ARC-C} &  \textbf{Winogrande} & \textbf{CommonSenseQA} \\
\midrule
\hline
\textbf{AQ-SGD 4}                       & 4           &  90.39	           &  93.30                  & 86.82                  & 86.56                                       \\ 
\textbf{AMAQ 4}                     & 4 ± 0.1     &  \textbf{91.74}      &  \textbf{93.73}         &  \textbf{91.23}	       & \textbf{86.89}      \\\hline
\midrule
\textbf{AQ-SGD 3}                       & 3           &  85.81               & 91.33                    & 77.03	               &  82.80              \\ 

\textbf{AMAQ 3}                     &  3 ± 0.1    & \textbf{85.87} 	    &  \textbf{92.44}         &  \textbf{85.79}        & \textbf{83.86}       \\\hline

\bottomrule
\end{tabular}}
\caption{Performance of Qwen 2.5 14B with All Layer Activation Quantization using AMAQ.}\label{all_layer}
\end{table*}

\begin{figure}[!ht]
    \centering
    \includegraphics[width=\linewidth]{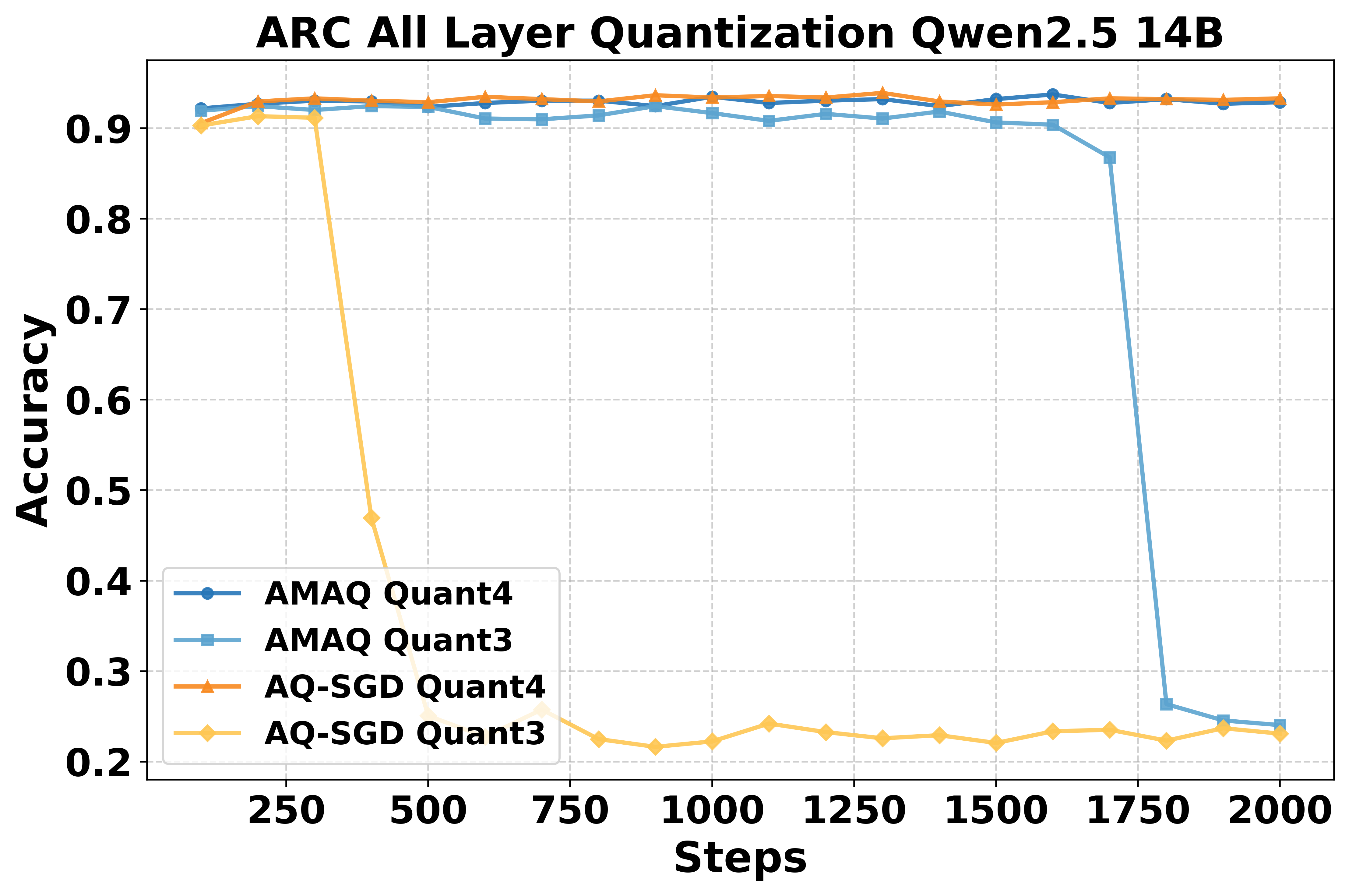}
     \caption{Performance and stability of Qwen2.5-14B on ARC under full-layer activation quantization.} \label{arc_all}
\end{figure}

\subsection{Stability Under Low-Bit Activation Quantization}
\begin{figure*}[!ht]
    \centering
    \includegraphics[width=\linewidth]{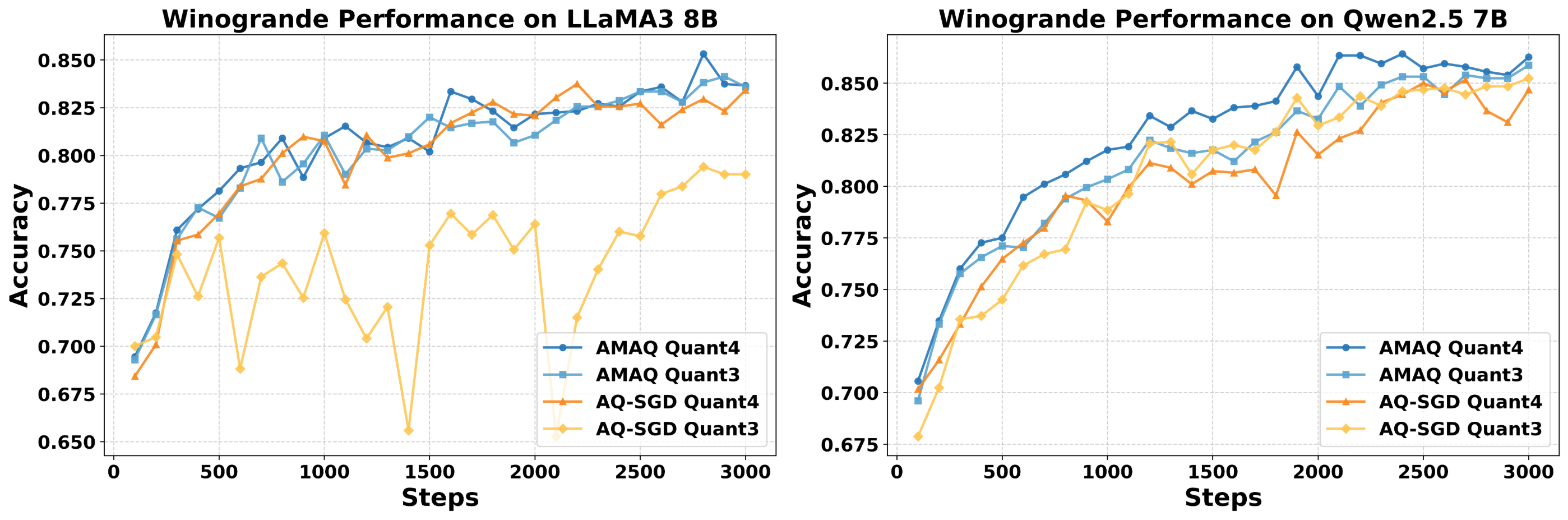}
     \caption{Performance and stability of LLaMA3 8B and Qwen2.5 7B on Winogrande under different bit-width for activation quantization.} \label{winogrando}
\end{figure*}

\begin{figure*}[!ht]
    \centering
    \includegraphics[width=\textwidth]{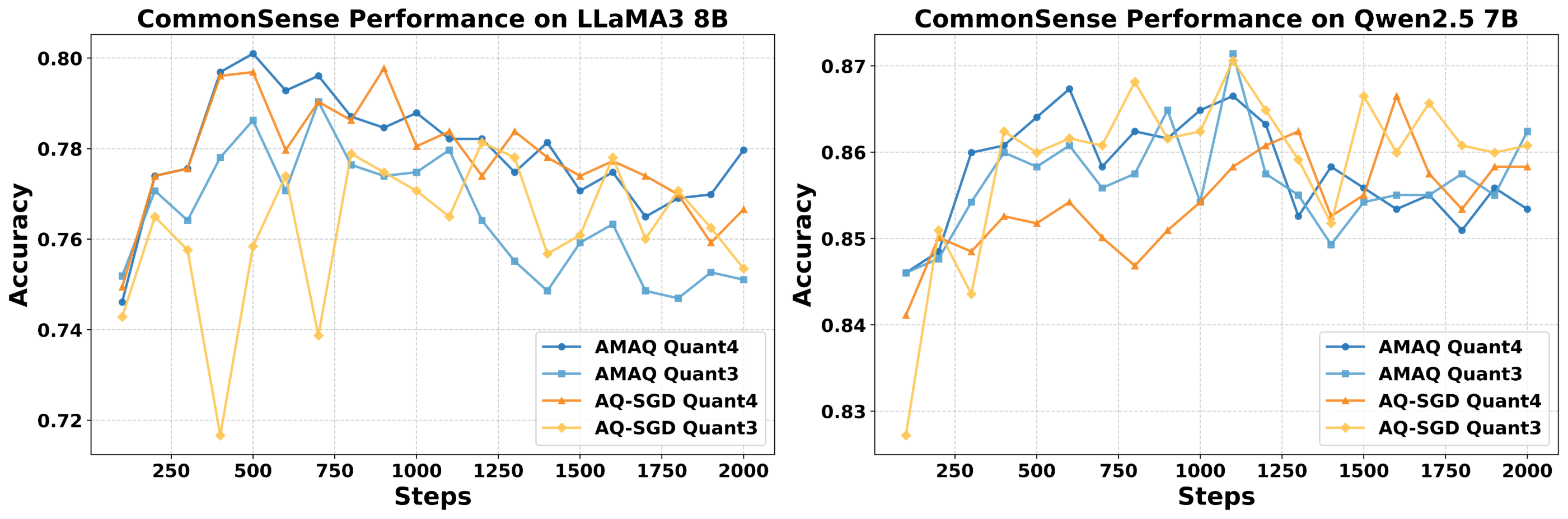}
     \caption{Performance and stability of LLaMA3 8B and Qwen2.5-7B on CommonsenseQA under different bit-width for activation quantization.} \label{commonsense}
\end{figure*}

There is a significant difference in training stability between 4-bit and 3-bit activation quantization. As shown in Figure~\ref{winogrando} and Figure~\ref{commonsense}, LLaMA3-8B maintains stable training under 4-bit quantization, but becomes highly unstable with 3-bit quantization, resulting in substantial performance degradation. However, applying AMAQ at 3 bits on LLaMA3-8B helps mitigate this issue, improving training stability. In addition, Qwen2.5-7B demonstrates greater robustness under low-bit quantization, showing better stability even at 3 bits.

Our findings suggest that stability is dependent not only on bit width but also on the model. Some models are more adaptable to low-bit activations than others. In general, reducing the activation precision to below 4 bits significantly increases the risk of unstable training.

Stability is also influenced by the number of layers subject to activation quantization. Applying low-bit quantization across all layers increases the likelihood of training collapse. Therefore, selectively applying activation quantization and balancing its scope across layers is important for maintaining stability. While AMAQ can not entirely prevent instability, it effectively slows down training collapse and enhances robustness in ultra-low-bit activation quantization, as demonstrated in Figure~\ref{boolq_all} and Figure~\ref{arc_all}.


\begin{figure*}[!ht]
    \centering
     \includegraphics[width=\textwidth]{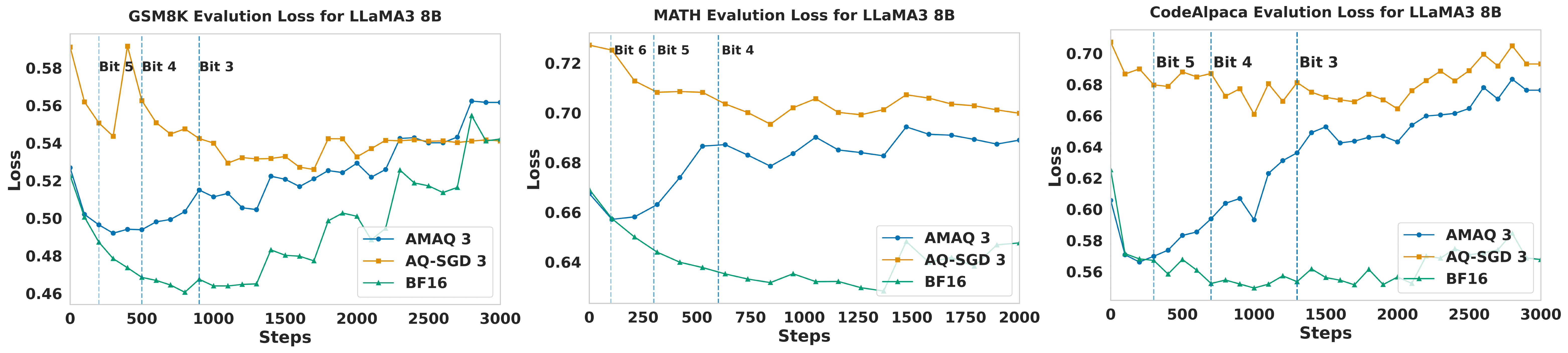}
     \caption{Evaluation loss on GSM8K, MATH, and CodeAlpaca benchmarks for AMAQ, comparing BF16 and AQ‑SGD across 3 bits for both input and output activation quantization on LLaMA3 8B.} \label{plot_Q3}
\end{figure*}

\begin{figure*}[!ht]
    \centering
     \includegraphics[width=\textwidth]{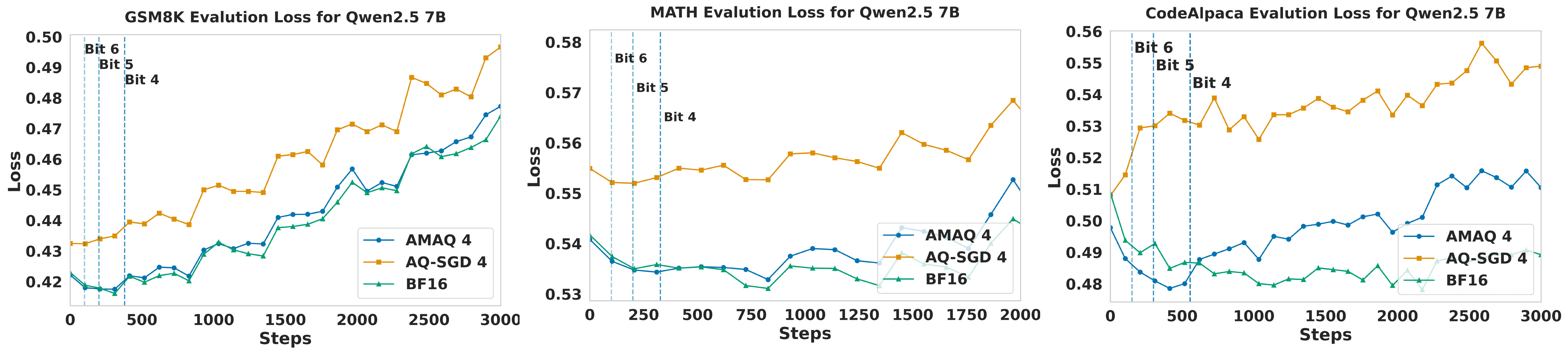}
     \caption{Evaluation loss on GSM8K, MATH, and CodeAlpaca benchmarks for AMAQ, comparing BF16 and AQ‑SGD across 4 bits for both input and output activation quantization on Qwen2.5 7B.} \label{plot_Qwen}
\end{figure*}

\end{document}